# Individualized Prediction of COVID-19 Adverse Outcomes with MLHO

Hossein Estiri,[1,2,3] Zachary H. Strasser,[1,3,4] Shawn N. Murphy[1,2,3,4,5]

[1] Laboratory of Computer Science, Massachusetts General Hospital, Boston, MA, 02144, USA.
[2] Research Information Science and Computing, Mass General Brigham, Somerville, MA, 02145, USA.
[3] Harvard Medical School, Boston, MA, 02115, USA.
[4] Department of Biomedical Informatics, Harvard Medical School, Boston, MA, 02115 USA.
[5] Department of Neurology, Massachusetts General Hospital, Boston, MA, 02114, USA.

## Abstract

The COVID-19 pandemic has devastated the world with health and economic wreckage. Precise estimates of adverse outcomes from COVID-19 could have led to better allocation of healthcare resources and more efficient targeted preventive measures, including insight into prioritizing how to best distribute a vaccination. We developed MLHO (pronounced as melo), an end-to-end Machine Learning framework that leverages iterative feature and algorithm selection to predict Health Outcomes. MLHO implements iterative sequential representation mining, and feature and model selection, for predicting patient-level risk of hospitalization, ICU admission, need for mechanical ventilation, and death. It bases this prediction on data from patients' past medical records (before their COVID-19 infection). MLHO's architecture enables a parallel and outcome-oriented model calibration, in which different statistical learning algorithms and vectors of features are simultaneously tested to improve prediction of health outcomes. Using clinical and demographic data from a large cohort of over 13,000 COVID-19-positive patients, we modeled the four adverse outcomes utilizing about 600 features representing patients' pre-COVID health records and demographics. The mean AUC ROC for mortality prediction was 0.91, while the prediction performance ranged between 0.80 and 0.81 for the ICU, hospitalization, and ventilation. We broadly describe the clusters of features that were utilized in modeling and their relative influence for predicting each outcome. Our results demonstrated that while demographic variables (namely age) are important predictors of adverse outcomes after a COVID-19 infection, the incorporation of the past clinical records are vital for a reliable prediction model. As the COVID-19 pandemic unfolds around the world, adaptable and interpretable machine learning frameworks (like MLHO) are crucial to improve our readiness for confronting the potential future waves of COVID-19, as well as other novel infectious diseases that may emerge.

# Introduction

The global spread of COVID-19, the disease caused by SARS-CoV-2, has resulted in the loss of around 700,000 lives. Repercussions of the pandemic have wreaked havoc on the economy, sending billions of people into lockdown and resulting in record-high unemployment around the world. The American Hospital Association estimates a total four-month financial impact of over $200 billion in losses for the U.S. healthcare system as a result of cancelled hospital services (e.g., cancelled non-elective surgeries and outpatient treatment) due to the COVID-19 pandemic.[1] Our inability to provide precise estimates of COVID-19 outcomes such as death, hospitalization, and need for ICU and ventilation, has contributed to lost opportunities for saving lives with personalized preventive measures and making smart resource allocation plans, such as the distribution of vaccinations. Although our inability to predict was partly due to the novelty of the disease, now the critical question is: do we have the data and technology to accurately predict outcomes?

Over the past decade, the U.S. federal government has made extensive investments to institute meaningful use of electronic health record (EHR) systems. Clinical data in EHRs, however, are still complex and have important quality issues, impeding their ability to address pressing health issues that require rapid responses. Nevertheless, biomedical researchers are increasingly applying data mining and machine learning techniques to clinical data for predicting health outcomes.

Recent studies have shown that COVID-19 disease severity and mortality are associated with a number of comorbidities including cardiovascular disease, diabetes mellitus, hypertension, chronic lung disease, cancer, chronic kidney disease, and obesity, and demographics including age, sex, and race/ethnicity.[2–8] A number of Machine Learning (ML) models have been developed to predict susceptibility in the general population, the likelihood of a positive diagnosis in a patient with symptoms, and prognosis in those with the disease.[9] Many of these models are based on a combination of demographics, comorbidities, symptoms, and biomarkers,[10–15] and use data from relatively small cohorts of COVID-19 patients. Jiangfent et al,[10] for example, performed a logistic regression analysis on 299 patients that identified age, lymphocyte count, lactate dehydrogenase, and oxygen saturation as independent predictors for mortality. Jiang et al,[15] used data from 53 patients to develop ML models that identified elevated alanine aminotransferase (ALT), the presence of myalgias, and an elevated hemoglobin as the most predictive features for disease severity, achieving approximately 70-80% accuracy in modeling COVID-19 morality. Huang et al[11] examined clinical data from 125 COVID-19 patients and identified the presence of comorbidities, increased respiratory rate, elevated C-reactive protein, and elevated lactate dehydrogenase as independently associated with a worse prognosis. While the inclusion of vital signs and biomarkers in these models may be highly predictive of some adverse outcomes of COVID-19 infection, they are typically measured after the patient has already started to show signs of the disease and may be at a point that is too late for a useful intervention.

Estiri et al. (2020) introduced the transitive Sequential Pattern Mining (tSPM) along with a dimensionality reduction algorithm (MSMR), and demonstrated that together, the two algorithms can successfully predict different health outcomes.[16,17] The goal in tSPM is to mine temporal data representations from clinical data for application in downstream ML. The MSMR algorithm – the short form stands for Minimize Sparsity, Maximize Relevance – applies high performance dimensionality reduction to tSPM representations. It takes the initial set of N features mined by the tSPM algorithm and provides a list of <400 features, through a 3-step process including frequency-based and information-based (mutual information and joint mutual information) filtering of a large number of features.[16,17]

We adapt the tSPM temporal representation mining and MSMR dimensionality reduction algorithms and make adjustments to architect an end-to-end ML framework that enables iterative feature and algorithm selection to predict Health Outcomes (MLHO, pronounced as melo). MLHO offers an architecture for parallel outcome-targeted calibration of the features and algorithms. The goal in MLHO is to mine relevant data representations and select the most efficient algorithmic solution for accurately predicting future health outcomes. As the COVID-19 pandemic unfolds in many areas of the world, being able to provide personalized predictions of adverse outcomes could be transformative to a healthcare system. In this study, we focus on predicting four outcomes (hospitalization, ICU admission, mechanical ventilation, and death) in patients with a verified COVID-19 infection, using their past medical records. Utilizing about 600 clinical features mined from patients' past medical records (before contracting COVID-19), we trained and tested predictive models that estimate risks of hospitalization, ICU admission, mechanical ventilation, and death. We also demonstrate and discuss the predictiveness of demographic features, such as age.

# Results

The overall mortality rate in the patient cohort was 5.3%. Table 1S provides a summary of demographic characteristics of this patient cohort. White patients had the highest mortality rate (7.3%) of any race or ethnic group. However, the rate of hospitalization and mechanical ventilation was highest among African American/black patients. Compared to females, male COVID-19 patients had a significantly higher chance of hospitalization, ICU admission, ventilation, and mortality. While the average age in the cohort of COVID-19 patients was 51, the average age of patients who were hospitalized, admitted to the ICU, or needed mechanical ventilation was between 62 and 64. The average age of mortality in COVID-19 patients was much higher at 78 (Table 1S).

Figure 1 summarizes the sequential scenarios in which the four adverse outcomes are observed in COVID-19 patients. Overall, in more than 72% of the patients, we did not observe any adverse outcomes. Approximately, 13% of patients were hospitalized and then discharged. The cumulative probability of patients needing to be admitted to the ICU was about 10%, which sorted in a declining order of subsequent sequential events of 5.4% (Hospitalized → ICU → Discharged), to 4.2% (Hospitalized → ICU → Ventilation), to less than 1% (Hospitalized → ICU → Died). This order supports our general hypothesis about the severity spectrum. Of the 5.3% who died, about 4% fall into the sequential scenarios of Hospitalized → ICU → Ventilation → Death, Hospitalized → ICU → Death, and Hospitalized → Death.

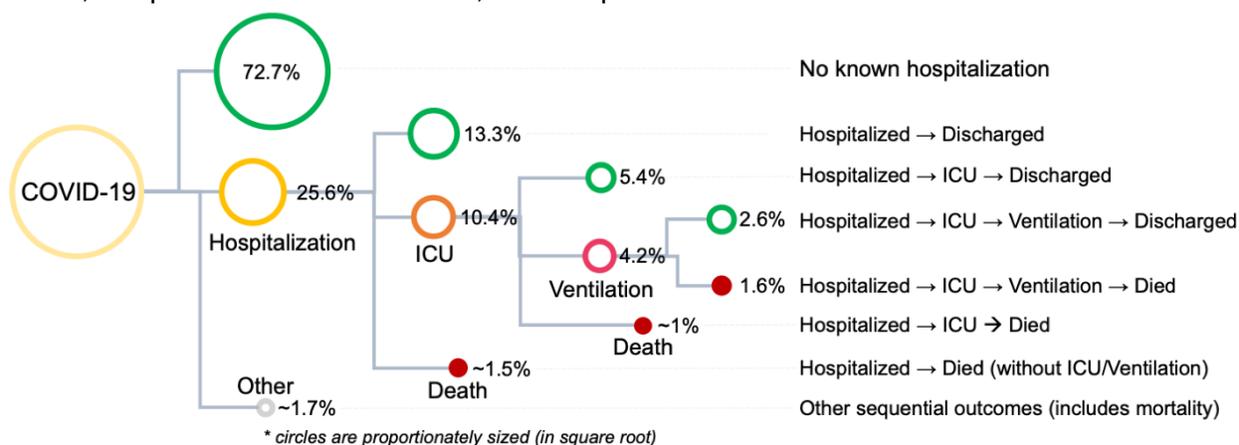

**Figure 1.** Probability of the sequential scenarios for outcomes after COVID-19 infection.

## Iterative feature and algorithm selection

In phase 1, MLHO mined over 60 thousand raw (e.g., diagnosis/medication/procedure codes) and 160 million transitive sequential (e.g., medication → diagnosis) representations. Through the iterative feature selection, these features were shrunk to over 2,300 representations, about 900 of which were raw features and about 1,400 were transitive sequential features. As described in the methods, both MSMR filter method and embedded methods (while training preliminary classification algorithms) were utilized in iterative sampling of train-test data. The other outcome of this step was identification of the top predictive algorithms. Figure 2 illustrates the Area Under the Receiver Operating Characteristic Curve (AUC ROC) results obtained from the 10 algorithms for estimating the risk of hospitalization. We found that two Boosting algorithms obtained the best overall results: The Stochastic Gradient Boosting (gbm) -- a.k.a., gradient boosting machine -- and the eXtreme Gradient Boosting (xgb) with DART booster (xgbDART).

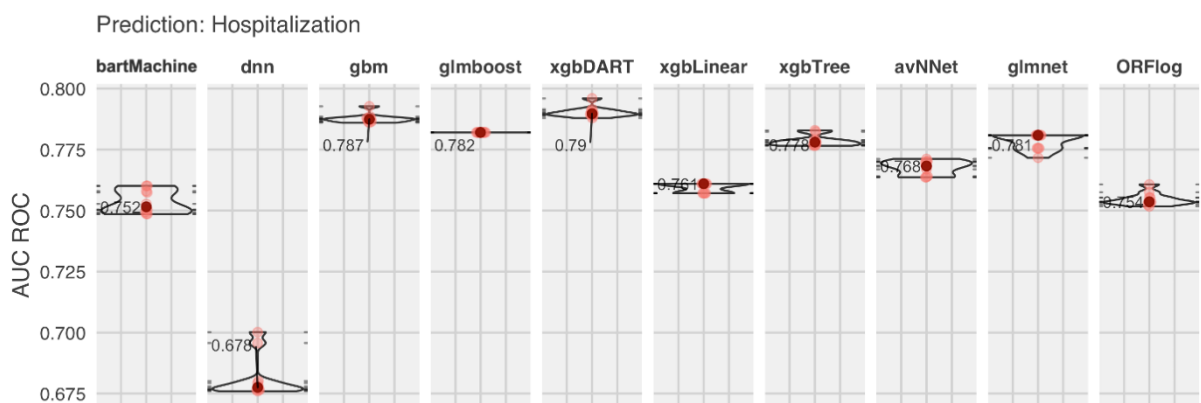

Algorithms from right to left: bartMachine: Bayesian Additive Regression Trees, dnn: Stacked AutoEncoder Deep Neural Network, gbm: Stochastic Gradient Boosting, glmboost: Boosted Generalized Linear Model, xgbDART: eXtreme Gradient Boosting (xgb) with DART booster, linear model solver (xgbLinear), tree learning (xgbTree), avNNet: model-averaged Neural Network, glmnet: Elastic-Net regularized generalized linear model, and ORFlog: Oblique random forest.

**Figure 2.** AUC ROC result of the preliminary classification (phase 1) for predicting hospitalization.

## Final modeling

We fed the shrunken clinical feature set (with ~2,300 features) to the two top algorithms for the final model training -- with 10 iterative train-test sampling and 5-fold cross-validation, which means for each outcome, we trained 20 final clinical models. For each outcome, we then trained 20 demographic models, in which we only used demographic features (age, gender, race, and ethnicity). We also trained 20 models (for each outcome) using the combination of demographic and clinical features. Table 1 presents the mean AUC ROC values obtained from the final models for each outcome and feature class, representing the model's discrimination power. Overall, the models that relied on both the combined demographic and clinical features resulted in the best modeling performance (as measured by AUC ROC). For predicting mortality and the need for ventilation, demographic models provided slightly superior discrimination power. In contrast, only using the clinical features resulted in superior performance in predicting ICU and hospital admission, over demographic-only models.

**Table 1.** The classification performance of the final models

|  | **Mortality** | **ICU** | **Ventilation** | **Hospitalization** |
|---|---|---|---|---|
| **Demographic** | 0.888* (0.887-0.888)** | 0.756 (0.751-0.76) | 0.773 (0.768-0.778) | 0.736 (0.735-0.737) |
| **Clinical** | 0.865 (0.862-0.868) | 0.784 (0.782-0.786) | 0.748 (0.741-0.755) | 0.777 (0.774-0.780) |
| **Demographic + Clinical** | **0.914 (0.910-0.918)** | **0.805 (0.801-0.810)** | **0.795 (0.791-0.797)** | **0.800 (0.798-0.800)** |

*Mean ** 95% Confidence Interval

To further evaluate the utility of demographic and clinical features for predicting adverse outcomes after COVID-19 infection, we also assessed the models' reliability for clinical interpretation using diagnostic reliability diagrams (calibration curves). To compare the models, we fitted smoothed trend lines as a representation of the overall calibration curve obtained from the 10 calibration plots for each model (Figure 3). The calibration plots are produced from the raw predicted probabilities computed by each algorithm (X axis) against the true probabilities of patients falling under probability bins (Y axis). In a well-calibrated model, the calibration curve appears along the main diagonal – the closer to the line, the more reliable. The calibration curves from the both the clinical and the combined demographic + clinical models provide more reliable predictions than the demographic-only models. Specifically, in predicting the need for ventilation, mortality, and ICU admission, only using the demographic features does not provide clinically reliable estimates of the risk for the adverse outcome.

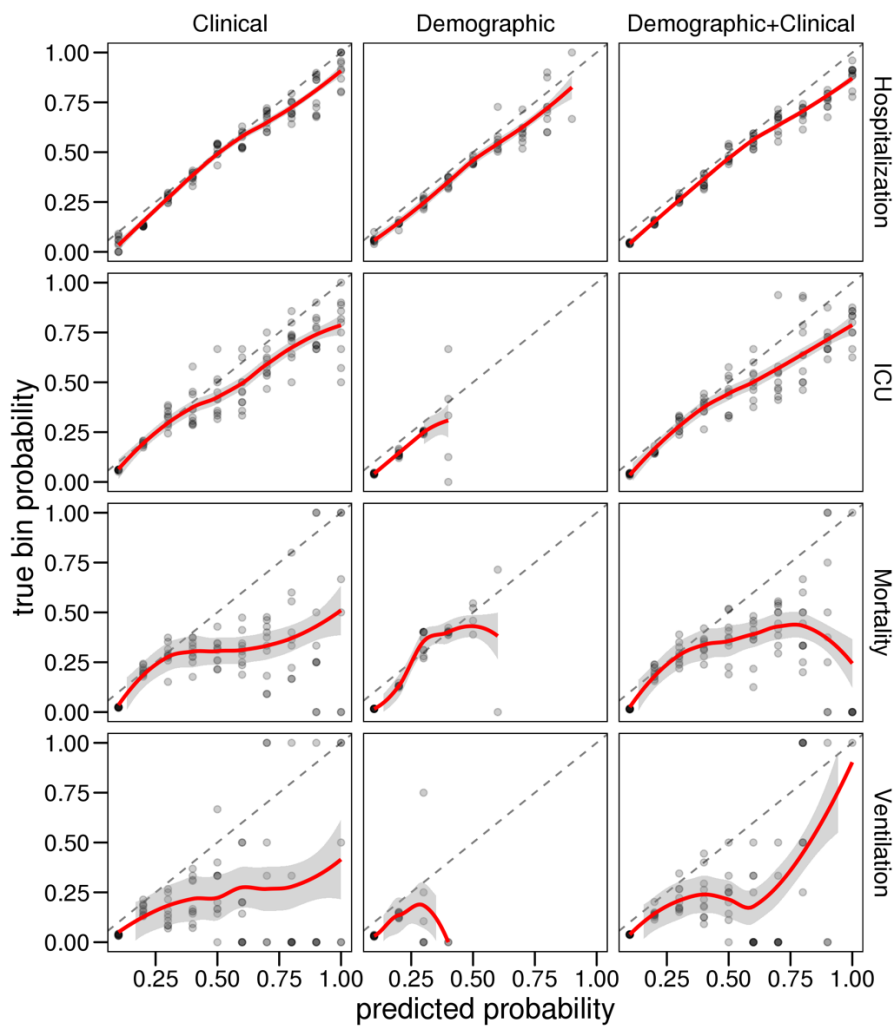

**Figure 3.** The diagnostic reliability diagrams (calibration curves).

## Features' relative influence

The features with the highest weights from the gbm models were assigned to 100, and the remaining values were scaled accordingly.[33] Using only the non-zero relative influence values, a final set of 598 features used for predicting at least one of the outcomes, were identified. Each feature was then assigned a label to assist with its interpretability. International Classification of Diseases (ICD) codes, both ICD-9 and ICD-10, were assigned to corresponding general comorbidities. This list included cardiovascular disease, chronic renal disease, chronic lung disease, neurological disorder, diabetes mellitus, and other chronic illness. The listed comorbidities were previously identified as associated with poor outcomes in COVID-19.[33] The "Other chronic illness" cluster includes comorbidities such as hypertension, hyperlipidemia, hypothyroidism, obstructive sleep apnea, and obesity. Other features that were included in the clinical model, but that did not fit neatly into a specific comorbidity, were assigned to clusters that described their general category. Many of these categories were labeled as either complex or general depending on the implied severity of the feature. For example, a previous billing code for a critical care encounter was interpreted as a "Complex Inpatient Episode", whereas a billing code for a 25-minute inpatient encounter was interpreted as a "General Inpatient Episode".

After transforming the features to more interpretable labels, the top five labels of each outcome (excluding age) are listed in Table 2. In each of the combined demographic + clinical models, age is the feature with the greatest influence on the outcome. However, after age, the features with influence varied significantly between the different models.

For mortality, the critical features were neurological disorder, cardiovascular disease, other chronic illness, diabetes mellitus and chronic kidney disease. The features with the greatest weight for predicting an ICU admission were healthcare utilization episodes including previous complex inpatient episodes, inpatient procedures, and inpatient medications. Some of the ICU features included transitive sequence features that captured common patient narratives. For example, a previous imaging episode followed by an inpatient medication was helpful for predicting an ICU admission. In this case, both previous complex inpatient episodes and inpatient medications were the most important features for determining the outcome. The critical features for predicting mechanical ventilation were a combination of chronic lung disease and significant healthcare utilization. Finally, the top features for a hospitalization were almost entirely billing codes that suggested a previous hospitalization.

**Table 2.** The top five ranked features from the clinical records that are associated with death, ICU admission, mechanical ventilation, and hospitalization.

| Death | ICU | Ventilation | Hospitalization |
|---|---|---|---|
| Neurological disorder *(Unspecified delirium)** | Imaging *(Electrocardiogram)-> Inpatient medication (Liquid acetaminophen)* | Complex inpatient episode *(Patient encounter billing, complex med)* | Inpatient episode *(Patient encounter billing)* |
| Cardiovascular disease *(Unspecified heart failure)* | Complex inpatient episode *(Patient encounter billing, complex med)* | Chronic lung disease *(Interstitial Pulmonary Disease)* | Complex inpatient episode *(Patient encounter billing, complex med)* |
| Other chronic illness *(Hypertension)* | Inpatient Procedure *(Venous blood draw)* | Inpatient medication *(Subcutaneous NPH insulin)* | Complex inpatient episode *(Patient encounter billing, complex med)* |
| Diabetes mellitus & Chronic renal disease *(Type II diabetes w/chronic kidney disease)* | Other chronic illness *(Hypertension)* -> Inpatient medication *(Liquid acetaminophen)* | Lab Value *(Proteinuria)* | Inpatient episode *(Patient encounter billing)* |
| Imaging *(CT scan of abdomen)* | Inpatient medication *(IV ceftriaxone)* | Diabetes mellitus *(Type II Diabetes w/Foot Ulcer)* | Inpatient procedure *(Chest X-ray)* |

*Parentheses below each of the interpreted terms include a shorthand version of the actual feature.

## Discussion

Using the MLHO framework, we developed models for predicting risks of hospitalization, ICU admission, need for mechanical ventilation, and death for patients infected with COVID-19. We were able to model the four adverse outcomes with about 600 features from patients' past medical records, before they contracted COVID-19. MLHO can leverage the past medical records in clinical repositories to quickly develop predictive models with acceptable accuracy. One could envision different applications for such predictions. For example, we could aggregate MLHO's predictions based on a population's expected rates of infection during a pandemic, to better allocate healthcare resources in preparation for a surge in cases. It could also help inform the allocation of critical care nurses and doctors, ventilators, hospital beds, and supplies in the case of a resurgence of the pandemic.

Being able to predict outcomes can create possibilities for better preventive measures. For instance, healthcare systems and regional health authorities can estimate the probability of an outcome, to identify patients who might be at higher risk and plan for more intensive preventive measures such as alerting the patients' primary care providers to prioritize healthcare maintenance visits. At such an appointment, one could ensure the patient is up-to-date on immunizations and have an in-person discussion on the importance of taking precaution against contracting COVID-19. Given there are limited resources for addressing this public health crisis, being able to estimate the relative risk even among otherwise seemingly similar populations could be extremely advantageous. For example, the algorithm could help determine who among a nursing home population has the greatest need for receiving a vaccination.

The average age of infection in this cohort was 51. The average age for hospitalization, mechanical ventilation, or ICU admission was between 62 and 64. But the average age of death was 78. While the best performing models were a combination of all of the clinical records including demographics and clinical features, age was always the highest weighted feature in every model for determining the adverse outcome. The models' dependence on age, emphasizes that COVID-19 is a disease that poses extreme risk to the elderly. No other features of the clinical record, including comorbidities or previous hospitalizations, have as much of an impact on predicting how an individual will respond to contracting COVID-19. Evaluating the top weighted features of the model excluding age, gives insight into how each of the models relies on specific sets of features for making its prediction. The most important features identified for predicting mortality, correspond well with many of the known associations of disease severity in COVID-19.[33] This includes cardiovascular disease, neurological disease, chronic kidney disease, diabetes mellitus, and other chronic illnesses. Features predictive of future ventilation include chronic lung disease and previous complex hospitalizations. For predicting ventilation, the model accurately identifies medically relevant features for who would need intensive respiratory support. In the case of ICU admissions and hospitalizations, the models depend to a greater extent on previous healthcare utilization such as inpatient stays, inpatient medications, and inpatient procedures. These healthcare utilization features are likely correlated with underlying disease, but they collectively predict a subsequent hospitalization or ICU admission better than the chronic diseases themselves. All four of the outcome models have their highest AUC when clinical records are incorporated with the demographics. The diagnostic labels in the clinical records are critical for predicting mortality and ventilation, whereas it is the healthcare processes embedded in the clinical records that are relatively more important for predicting ICU and hospital admissions.

# Method

MLHO mines both sequential temporal and raw data representations from clinical data and performs iterative feature and algorithm selection in a 2-phase process to learn from the unfolding COVID-19 pandemic (Figure 4). MLHO's architecture enables a parallel outcome-targeted calibration of the features and algorithms, in which different statistical learning algorithms and vectors of features are simultaneously tested and leveraged to improve prediction of health outcomes. The four adverse outcomes of interest in this study reflect a hypothetical sequential spectrum of outcome severity in patients with verified COVID-19 infection, ranging from hospitalization to ICU admission, to need for mechanical ventilation, and ultimately to death (Hospitalization → ICU → Ventilation → Death).

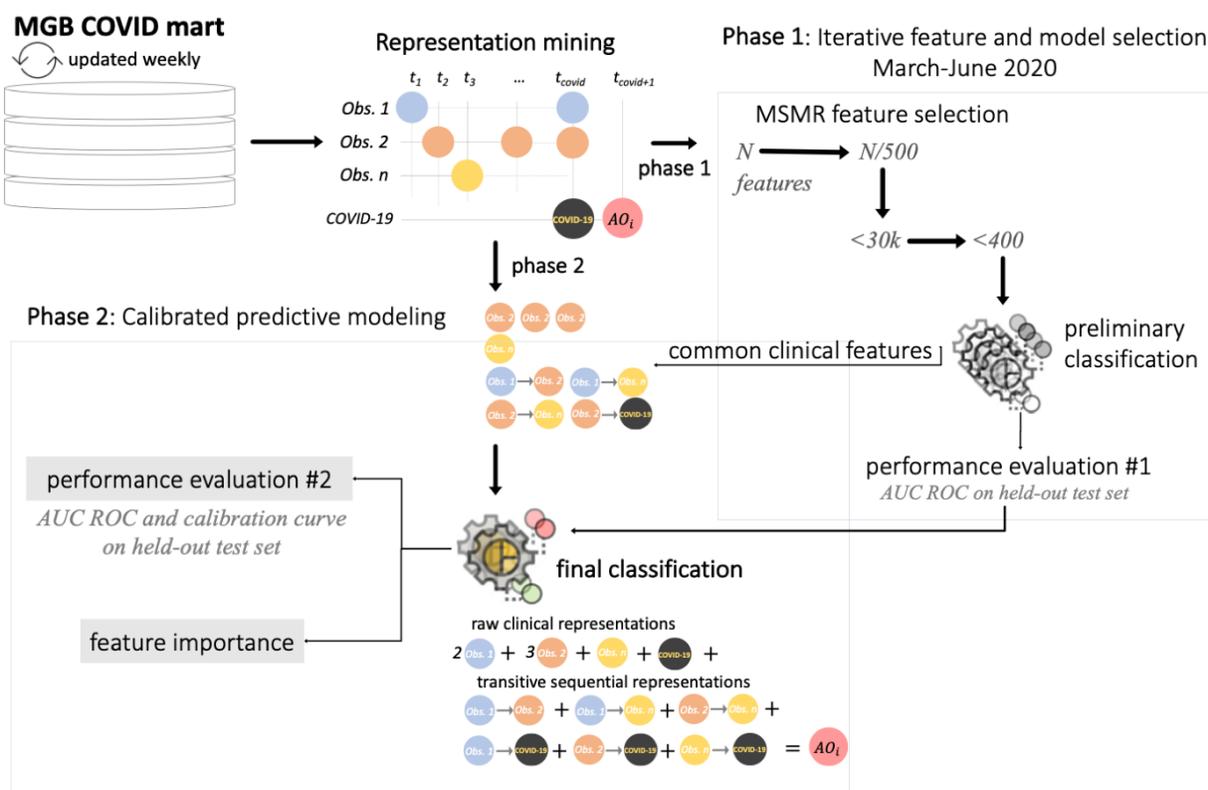

**Figure 4.** Implementation of the MLHO framework for predicting COVID-19 adverse outcomes

# Code availability

The computer code for MLHO is available at: https://hestiri.github.io/mlho

# Experimental Setup

We used electronic health records data from over 13,400 patients with a confirmed case for COVID-19 between March and September 2020 and who had at least 1 year of medical history with Mass General Brigham (MGB), since 2016. Table 1S presents general demographic information about the study cohort. For each patient, we only included clinical records from 14 days prior to the positive COVID-19 test date. This temporal buffer ensures that no COVID-19-related medical conditions are included in the model as a risk factor. The use of data for this study was approved by the Mass General Brigham Institutional Review Board (2020P001063) – a waiver of informed consent was also approved. All research was performed in accordance with relevant guidelines and regulations. We randomly split the data with an 80-20 training-to-testing ratio. We iterated the train-test sampling 10 times to account for possible patient population differences caused by the sampling.

## Representation mining

We used the tSPM algorithm to mine temporal sequential representations. Given a list of clinical records $R_1, R_2, \ldots, R_n$ for patient $p$ at times $t_{i1} \leq t_{i2} \leq \ldots \leq t_{ik_i}$, the tSPM algorithm mines all avector of transitive sequential patterns $X_{ij}$ from possible pairs of distinct records $(R_i, R_j)$, where $i \neq j \leq n$, by setting $r_{ij}$ (as samples of random variable $X_{ij}$ for patient $p$) to be 1 if $k_i \geq 1, k_j \geq 1$ and $t_{i1} \leq t_{ji}$, and 0 otherwise. We also mined raw representations from the clinical data, which are composed of all clinical records. To obtain a count of the raw records, for each patient $p$, we sum the frequency of $k_1, k_2, \ldots, k_n$, as samples of a random variable $X_i$. On the training sets, we performed feature and algorithm selection (phase 1) and the final predictive modeling (phase 2). We feed the combined representations $X' = (X_i \cup X_{ij})_{i \neq j}$ to the feature selection step.

## Phase 1: Iterative feature and algorithm selection

In phase 1, as many aspects of the COVID-19 pandemic was still unknown, we perform iterative feature and model selection on weekly-updated data from COVID-19 patients during the first four months of the pandemic (March-June 2020). Unlike other studies that begin with a limited set of hypothetical risk factors, we took a primarily inductive approach to selecting clinical representations for predicting the risk of adverse outcomes in COVID-19 patients. First, we apply a filter method for feature selection, using a computational algorithm that minimizes sparsity and maximizes relevance (MSMR[16,17]). Step 1 in MSMR is to cut the initial combined vector of representations $X'$ to those that were observed in fewer than 0.2 percent of the patients. On the remaining representations, step 2 in MSMR computes the mutual information with the outcome variable $Y = (y_1, \ldots, y_i)$, which in this study includes labels for hospitalization, ICU, ventilation, and death. Mutual information,[18,19] in this case, measures the amount of information that each remaining representation contains about the outcome. Given the joint probability distribution $P_{X'Y}(x', y)$, the mutual information between them is denoted as $I(X':Y)$ is:

$$\sum_{x'y} P_{X'Y}(x', y) \times \log \frac{P_{X'Y}(x',y)}{P_{X'}(x') \times P_Y(y)} \qquad (1)$$

We cut the remaining representations from $X'$ by ranking based on the mutual information coefficient, and update $X'$ to a list of around 30,000 representations with the highest mutual information. In the third and final step, MSMR computes the joint mutual information (JMI)[20] score for the updated vector of remaining representations, $X'$. The algorithm starts with a set $S$ containing the top feature according to mutual information, then iteratively adds to $S$ the features maximizing the joint mutual information score

$$J_{jmi}(X') = \sum_{X'^* \in S} I(X'X'^*; Y) \qquad (2)$$

Where the random variable $X'X'^*$ corresponds to the joint distribution of $X'$ and $X'^*$. As a result, JMI also takes into account the redundancy between the features – i.e., reducing multicollinearity among covariates. Second, we combine the feature selection with a preliminary evaluation of algorithms for the prediction task.

Using the <400 clinical features identified by the MSMR algorithm, we train a set of preliminary classification algorithms that perform embedded feature selection. The preliminary classification

Estiri et al.serves two goals. First, we screened features used in those algorithms to compile a list of the common features used for modeling the 4 outcomes. Second, during each sampling iteration, we computed the Area Under the Receiver Operating Characteristic Curve (AUC ROC) on the held-out test sets to evaluate the algorithms' performance for predicting the outcome labels. We used 10-fold cross-validation to train the prediction algorithm -- therefore, a 72-8-20, train-evaluation-test split. We tested 10 classification algorithms -- bartMachine: Bayesian Additive Regression Trees,[21,22] Stacked AutoEncoder Deep Neural Network (dnn), Stochastic Gradient Boosting (gbm),[23,24] glmboost: Boosted Generalized Linear Model, eXtreme Gradient Boosting (xgb)[25] with DART booster (xgbDART),[26] linear model solver (xgbLinear), tree learning (xgbTree), model-averaged Neural Network (avNNet), Elastic-Net regularized generalized linear model (glmnet),[27,28] and Oblique random forest (ORFlog).[29]

## Phase 2: Calibrated predictive modeling

The iterative feature and algorithm (preliminary classification) selection in phase 1 results in a set of common features and a ranking of classification algorithms. The predictive modeling in phase 2 are therefore calibrated specific outcomes. We use the top-performing algorithms (as measured by AUC ROC) and the common clinical features to perform a final round of training predictive models with 10 train-test sampling iteration and 10-fold cross-validation in phase 2. We summarize and report the AUC ROCs and calibration curves from the final models as well as the features' relative influence values. Further, we evaluate the role of demographic covariates in predicting the adverse outcomes after COVID-19 infection. To do so, we run 3 classes of models using: 1) only clinical features (from MLHO's phase 1), 2) only demographic features (age, gender, race, and ethnicity), and 3) clinical plus demographic features.

As we shall reveal in the results, the gbm algorithm was one of the top 2 performing algorithms. As a result, we computed the relative feature influence metrics from the final gbm models to measure features' importance. In the gbm model, boosting estimates $\hat{F}(X)$ as an 'additive' expansion of the form

$$\hat{F}(X) = \sum_{m=0}^{M} \beta_m h(X; a_m) \quad (3)$$

Where the expansion coefficients $\{\beta_m\}_0^M$ and the base learner's -- $h(X; a)$-- parameters $\{a_m\}_0^M$ are jointly fit to the training data in a feedforward process. A regression tree model specializes tue base learner $T_m(X; \{R_{jm}\}_1^J)$ partitions the feature space into $J$ disjoint regions $\{R_{jm}\}_{j=1}^J$ to predict a different constant value for each.[30] The relative influence (or contribution), $I_j^2$ averages the improvement made by each variable when it is permuted from all trees in which the given variable was incorporated.[30,31] In an additive tree model, the relative influence measure is provided by Friedman and Meulman (2003)

$$I_j^2 = \frac{1}{M} \sum_{m=1}^{M} I_j^2(T_m) \quad (4)$$

where $I_j^2(T)$ is the measure of relevance for a single tree $T$.[32]